\documentclass[letterpaper, 10 pt, conference]{ieeeconf}  

\IEEEoverridecommandlockouts                              

\usepackage{amsmath} 
\usepackage{amssymb}  
\usepackage{mathtools}
\usepackage{multirow}
\usepackage{cite}
\usepackage{graphicx}
\DeclarePairedDelimiter{\norm}{\lVert}{\rVert} 
\usepackage{xcolor}
\usepackage{hyperref}

\title{\LARGE \bf
Human Balance Assistance during Standing with a Floating-base Manipulator }
\title{\LARGE \bf
Improving Standing Balance Performance through the Assistance of a Mobile Collaborative Robot}

\author{Francisco J. Ruiz-Ruiz$^{1,+}$, Alberto Giammarino$^{2,+}$, Marta Lorenzini$^{2}$, Juan M. Gandarias$^{2}$,\\ Jes\'us M. G\'omez-de-Gabriel$^{1}$ and Arash Ajoudani$^{2}$
\thanks{This work was supported in part by the ERC-StG Ergo-Lean (Grant Agreement No.850932), in part by the European Union’s Horizon 2020 research and innovation programme under Grant Agreement No. 871237 (SOPHIA), in part by the Spanish project UMA CEIATECH-23 and the University of Málaga.}
\thanks{$^{1}$ Francisco J. Ruiz-Ruiz and Jes\'us M. G\'omez-de-Gabriel are with the Robotics and Mechatronics Group, University of Málaga, 29071 Málaga, Spain
        {\tt\small \{fjruiz2, jesus.gomez\}@uma.es}}%
\thanks{$^{2}$ Alberto Giammarino, Marta Lorenzini, Juan M. Gandarias, and Arash Ajoudani are with the HRI$^{2}$ Lab, Istituto Italiano di Tecnologia, Genoa, Italy.
        {\tt\small \{alberto.giammarino, marta.lorenzini, juan.gandarias, arash.ajoudani\}@iit.it}}
\thanks{$^{+}$ Contributed equally to this work.}
}

\begin{document}

\maketitle
\thispagestyle{empty}
\pagestyle{empty}

\begin{abstract}

This paper presents the design and development of a robotic system to give physical assistance to the elderly or people with neurological disorders such as Ataxia or Parkinson's. In particular, we propose using a mobile collaborative robot with an interaction-assistive whole-body interface to help people unable to maintain balance. The robotic system consists of an Omni-directional mobile base, a high-payload robotic arm, and an admittance-type interface acting as a support handle while measuring human-sourced interaction forces. The postural balance of the human body is estimated through the projection of the body Center of Mass (CoM) to the support polygon (SP) representing the quasi-static Center of Pressure (CoP). In response to the interaction forces and the tracking of the human posture, the robot can create assistive forces to restore balance in case of its loss. Otherwise, during normal stance or walking, it will follow the user with minimum/no opposing forces through the generation of coupled arm and base movements. As the balance-restoring strategy, we propose two strategies and evaluate them in a laboratory setting on healthy human participants. Quantitative and qualitative results of a 12-subjects experiment are then illustrated and discussed, comparing the performances of the two strategies and the overall system.
\end{abstract}

\section{Introduction}
\label{sec:introduction}

Neurological disorders such as Cerebellar Ataxia or Parkinson's usually influence or even obstruct the individuals' capability to maintain balance. Ataxic patients, for instance, lack limb coordination, which varies in severity depending on the stage of the disease \cite{martino2014}. Due to their poor joint coordination, ataxic patients develop atypical gait cycles and impaired balance that make them prone to falling.

\begin{figure}
    \centering
    \includegraphics[width=0.7\columnwidth]{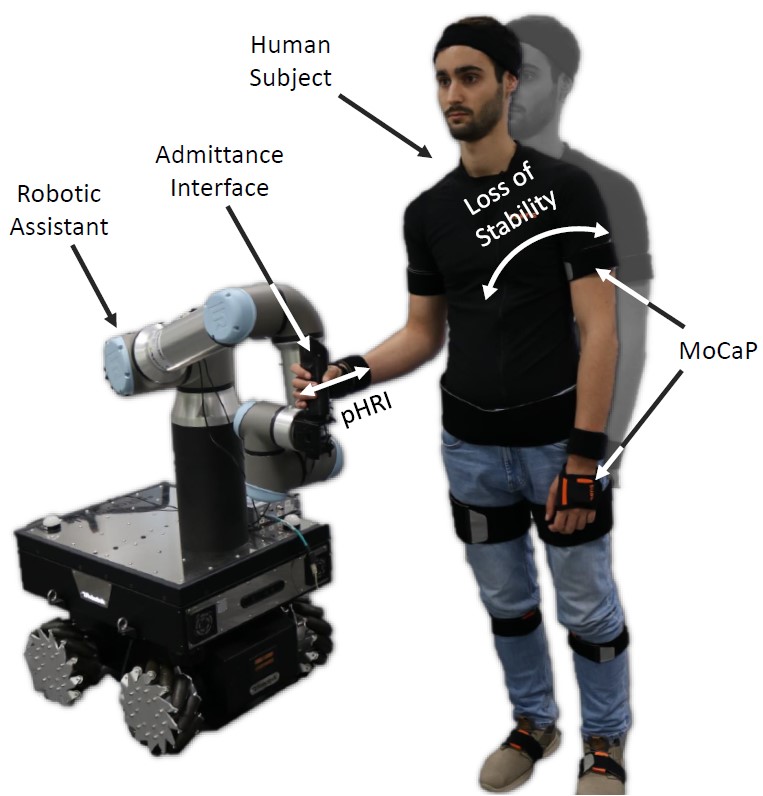}
    \vspace{-4 mm}
    \caption{A participant being assisted by a mobile robotic manipulator  when loss of balance is detected through a Motion Capture (MoCap) system. The robotic platform implements an assistive strategy based on the person's Center of Mass (CoM) and Support Polygon (SP) to provide support when needed.}
    \vspace{-0.5cm}
    \label{fig:cover_figure}
\end{figure}

On the other hand, due to the recent advances in physical Human-Robot Interaction (pHRI) \cite{RuizR2021,gandariasIros,kim2021human, skilltransfer}, and power augmentation~\cite{RAISAMO2019131}, the use of wearable and collaborative robotic systems for walking assistance is gaining momentum.
Among these technologies, exoskeletons are one of the most common devices to address the challenge of balance assistance,
\cite{Farkhatdinov2019, zhang2018}. 
Despite their high potential, most existing exoskeletons present fewer degrees of freedom (DoFs) than the human limbs, hence they tend to hinder the users' motion. In addition, the added device inertia usually makes the gait (even when the person is able to maintain balance) appear more static.

Supernumerary limbs as an alternative category offer a less obtrusive solution that can deal with this limitation. For instance, a supernumerary robotic system for gait assistance consisting of a pair of retractable legs is presented in~\cite{parietti2015}. 
The same device was later used to assist users with sitting and standing~\cite{parietti2017}.
Maekawa \textit{et al.} designed a wearable robotic tail built as a mass fixed to the tip of an aluminum rod actuated by a high torque DC motor attached to the human torso~\cite{tailMaekawa2020}. 
Concerning the challenge addressed in this paper, this technology is more suitable than the use of exoskeletons. However, the unexpected movement of mechanical limbs may feel unnatural for the human user. Besides, the weight of the device is supported entirely by the user \cite{srlReview}.

Alternative solutions to exoskeletons and supernumerary limbs have also been developed. 
A wearable backpack-like system capable of exerting recovery forces based on an actuated gimbal mechanism and a flywheel was proposed in~\cite{flywheelImplementation}.
The prototype was later tested on humans with promising results~\cite{vallery2020}. Nevertheless, the main limitation of this prototype is its low maximum payload, which may make it impractical for long-term use.
A robotic cane-like system with a 2 DoFs wheel is presented in~\cite{vanlamCane}. The wheel allows the cane to move along the X and Y axes, but the rotation around the Z-axis is not considered, constraining user mobility.

A potential alternative to the solutions presented above would involve an independent robotic system that acts as a supernumerary body~\cite{mocaman} capable of applying external forces to the patient, emulating the assistance that another person (i.e., a therapist) would provide. In this regard, collaborative mobile manipulators seem to be a workable and promising solution. However, this approach has hardly been addressed in the literature.
To the best of our knowledge, the only work that uses a mobile manipulator for this purpose was presented by Xing \textit{et al.} in~\cite{XING2021102497}. The proposed system uses a mobile manipulator as a cane to help the elderly during walking. 
This solution involves a robotic system that 
provides vertical support when moving the end-effector under a certain pre-defined threshold. However, the robot does not have any information about the human state, which may lead to an inadequate assistance.

This work aims to get the maximum benefit from the physical assistance that a mobile supernumerary robotic body~\cite{mocaman} could provide to a person when the balance is disturbed or lost (see Fig.~\ref{fig:cover_figure}). Hence, 
this paper proposes a methodology to detect and compensate for unbalancing situations under standing conditions based on the user state. In particular, the contributions of this paper are the following:
\begin{itemize}
	\item We present a method to detect the risk of falling based on the user's kinodynamic states, and integrate it in the controller of a mobile supernumerary manipulator to provide physical assistance when a loss of balance is detected (see Fig.~\ref{fig:cover_figure}). 
	\item We propose a first strategy that provides fixed physical support to the user. The end-effector behaves like a mass-spring-damper system, with a reference pose placed at the last end-effector pose when the human was under a balance condition, giving fixed compliant support.
	\item We also propose a second strategy that, unlike the first, provides variable physical assistance. Here, the robot applies a compliant and variable force to the human arm depending on the unbalance status, giving a compliant balance compensation.
	\item We conduct an in-lab experimental evaluation to compare the proposed strategies' performance with the most related strategy of the current state-of-the-art. 
\end{itemize}
A motion capture (MoCap) system monitors the human state, namely the Center of Pressure (CoP) and the Support Polygon (SP). 
Due to the difficulties of measuring the CoP, it is estimated by projecting the Center of Mass (CoM) over the ground plane.
An unbalancing situation is detected when the CoP is outside of a pre-defined region contained in the SP. Then, an Omni-directional mobile robotic manipulator governed by a whole-body controller in conjunction with an admittance controller gives physical support according to the aforementioned strategies. An experimental session with 12 healthy human subjects is performed. During the experiments, the proposed strategies are tested and compared along with the strategy proposed in~\cite{XING2021102497}, considered a benchmark. A discussion based on a quantitative and qualitative comparison of the three approaches is carried out considering the data recorded during the experiments and the questionnaires filled in by the participants.

\section{System Overview}
\label{sec:systemOverview}

\subsection{Robotic Platform}
\label{subsec:robotic_platform}
In this work, the robotic platform Kairos (Fig.~\ref{fig:cover_figure}) is used as a robotic assistant.
It consists of a Robotnik SUMMIT-XL STEEL mobile platform, an Omni-directional mobile base, and a high-payload (16kg) 6-DoFs Universal Robot UR16e arm attached on top of the base. In addition, a physical admittance-type interface is developed and added at the manipulator's end-effector to facilitate assistance to the user. A handle forms the interface that the human can grasp, and a force-torque (F/T) sensor measures the wrenches applied on the handle.

\subsection{Control Framework}

\begin{figure}
    \centering
    \includegraphics[width=0.7\columnwidth]{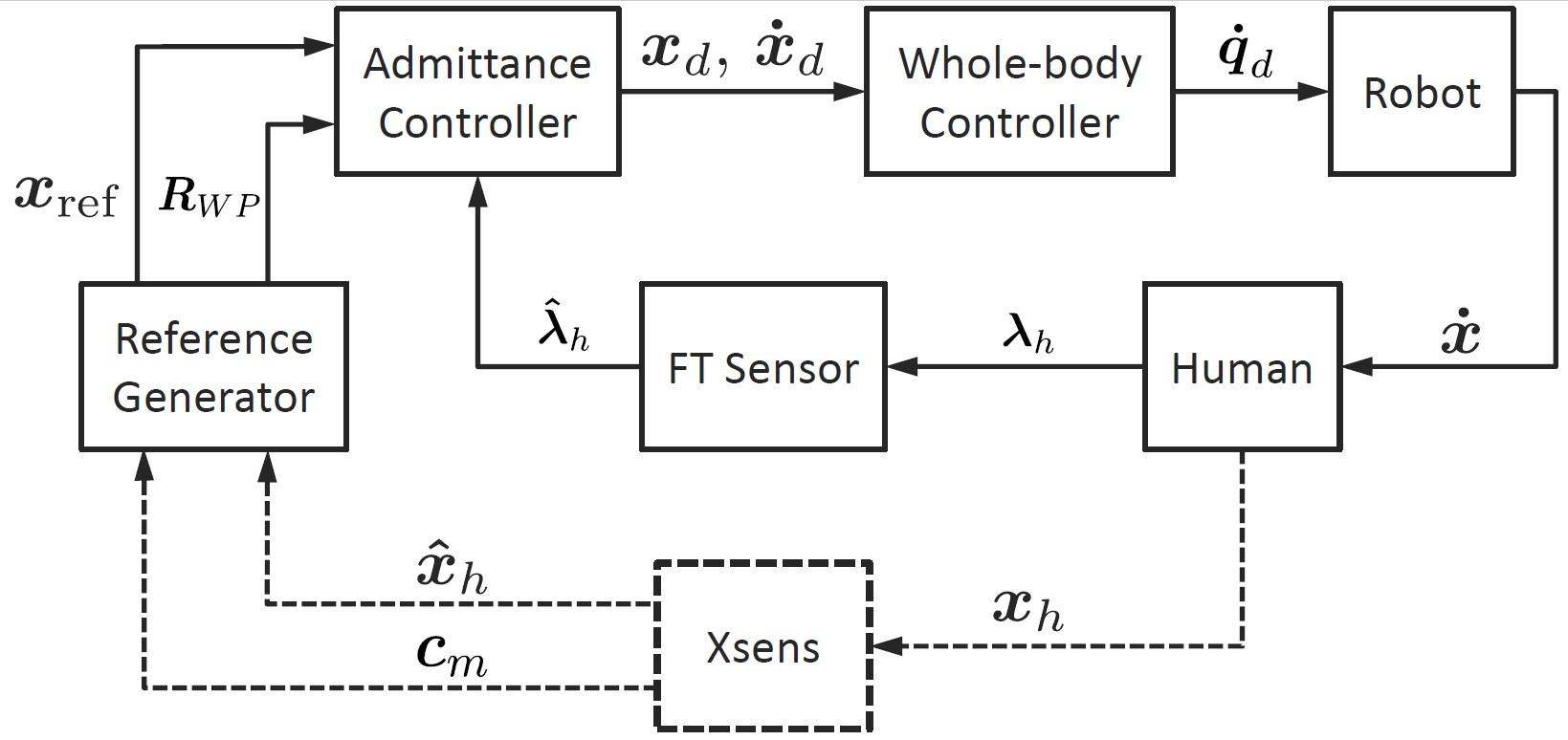}
    \caption{Block diagram of the 
    control framework (solid lines) and additional feedback needed for implementing the assistive actions (dashed lines)}
    \vspace{-0.5cm}
    \label{fig:block_diagram}
\end{figure}

The framework developed in this work is represented in the block diagram of Fig.~\ref{fig:block_diagram}, where the controller implemented on the robot and the human feedback are drawn respectively in solid and dashed lines. The control architecture has three main components: the reference generator, the admittance controller, and the whole-body controller. These components and their interaction are explained in detail below. 

Based on the current human state, the reference generator computes and sends a reference pose
$\boldsymbol{x}_{\textrm{ref}} \in \mathbb{R}^6$ and the rotation matrix from principal to world frame $\boldsymbol{R}_{WP} \in \mathbb{R}^{3 \times 3}$ (explained in detail in Sec.~\ref{subsec:unstable_state}) to the admittance controller, that implements the following control law expressed in the world frame:
\begin{equation}
\label{eq:admittance_controller}
\small
    \boldsymbol{X}_d(s) = \frac{\boldsymbol{\hat{\Lambda}}_h(s)+\boldsymbol{K}_{adm}\boldsymbol{X}_{ref}(s)}{\boldsymbol{M}_{adm}s^2+\boldsymbol{D}_{adm}s+\boldsymbol{K}_{adm}},
\end{equation}
where $\boldsymbol{M}_{\textrm{adm}}$, $\boldsymbol{D}_{\textrm{adm}}$ and $\boldsymbol{K}_{\textrm{adm}} \in \mathbb{R}^{6\times 6}$ are the desired mass, damping and stiffness matrices, $s$ is the Laplace variable, and $\boldsymbol{X}_d(s)$, $\boldsymbol{X}_{ref}(s)$ and $\boldsymbol{\hat{\Lambda}}_h(s)$ are the Laplace transforms of the desired pose $\boldsymbol{x}_d \in \mathbb{R}^6$, reference pose $\boldsymbol{x}_{ref}$ and human measured wrench $\boldsymbol{\hat{\lambda}}_h \in \mathbb{R}^6$, respectively.

The whole-body controller finds the whole-body joint velocities $\boldsymbol{\dot{q}}_d \in \mathbb{R}^9$ resulting in the desired motion at the end-effector ($\boldsymbol{\dot{x}}_d$ and $\boldsymbol{x}_d$) solving a Hierarchical Quadratic Programming (HQP) problem composed of two tasks. At higher priority, the closed loop inverse kinematics (CLIK) problem is solved~\cite{seraji1990improved}. Then, the cost function at lower priority exploits the movement of the mobile base to keep the arm close to a preferred configuration~\cite{nakanishi2005comparative, wu2021unified}. Read \cite{giammarino2022super} for more details.

\subsection{MoCap System}
The MoCap system used in this study is the Xsens (Fig.~\ref{fig:cover_figure}). It consists of 17 Inertial Measurement Units (IMUs) worn by the human. This system is employed to get real-time feedback from the human status in terms of body configuration ($\boldsymbol{\hat{x}}_h$) and CoM ($\boldsymbol{c}_m$) (dashed lines in Fig.~\ref{fig:block_diagram}). This information is then used by the reference generator and the admittance controller to implement the assistive strategies, as explained in the following section.

\section{Methodology}
\label{sec:methods}
\begin{figure*}
    \centering
    \includegraphics[width=0.65\textwidth]{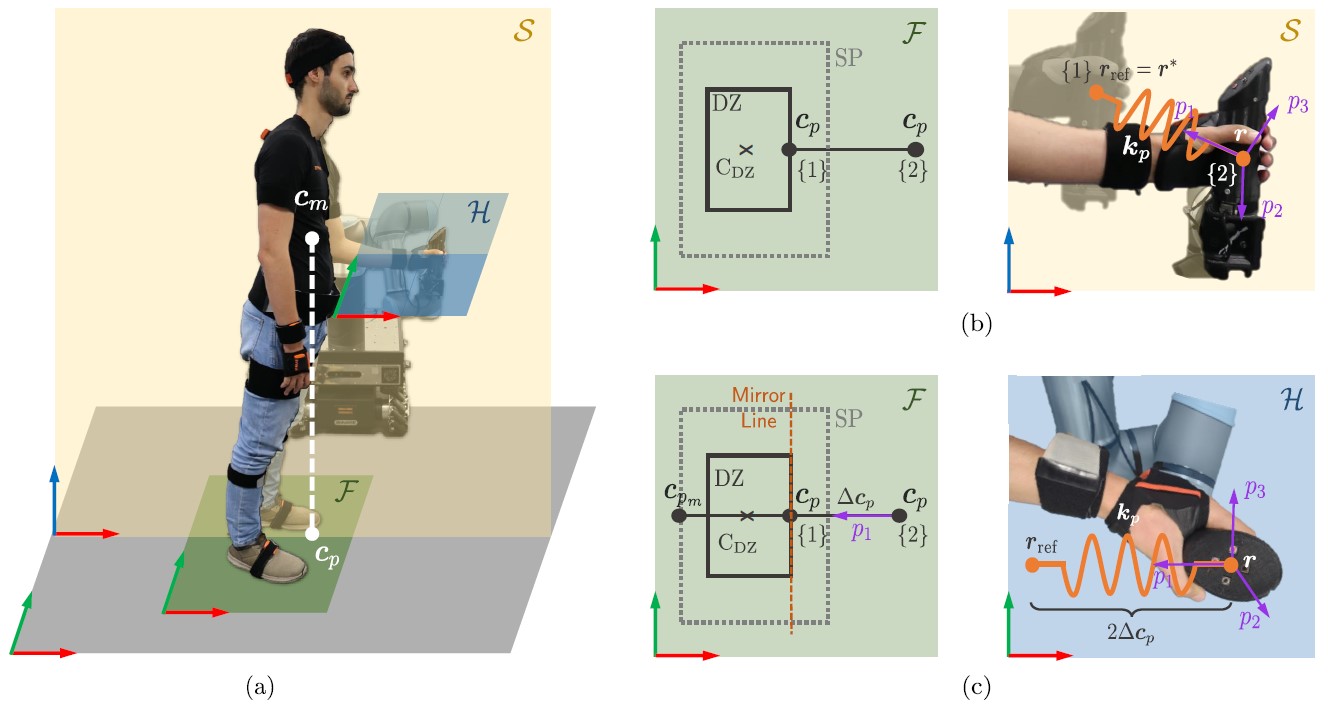}
        \vspace{-5 mm} 
    \caption{Illustration of the proposed strategies. (a) A participant using the robot as an aid to recover the balance is depicted, along with the sagittal plane $\mathcal{S}$, the hand plane $\mathcal{H}$ and the feet plane $\mathcal{F}$. (b), (c) A graphical explanation of the rationale behind FSA and MBA strategies in the planes of interest, respectively. The XYZ-RGB convention is used in all the reference frames.}
    \vspace{-0.5cm}
    \label{fig:methods}
\end{figure*}
\subsection{Assumptions}
The human balancing mechanism is a complex process that can be inferred by various parameters. There are several indexes to quantify the balancing state associated with the level of a neurological disease (e.g., the anatomical joint angles~\cite{index1} and the time-varying multi-muscle co-activation function~\cite{index2}). Nevertheless, one of the indexes that provide more straightforward information about the equilibrium of a person is the CoP~\cite{cop1, cop2}. 
Humans act over their muscles to alter the velocity and position of their CoM, thus changing their CoP. When the CoP is inside the SP, the human body is in static equilibrium.

In this work, we assume that the CoP is the projection of the CoM on the ground plane. This is true under certain conditions, for instance, when the human is not interacting with the external environment and performs quasi-static movements. This assumption is reasonable for the application addressed since the robot should only assist the humans if they cannot recover the balance without external support. 

Theoretically, a healthy human should keep the equilibrium when the CoP is inside the SP, but a margin is considered in our approach for safety reasons. We assume that if the CoP is inside a Deadband Zone (DZ) region, the human can recover the balance without external help. This region should be customized for each participant since each person possesses different balancing capabilities. Thus,
it changes from person to person based on a procedure explained in Sec.~\ref{subsec:experimental_protocol}. The two strategies are described below, distinguishing between the stable (i.e., CoP inside DZ) and unstable (i.e., CoP outside DZ) states.

\subsection{Stable State: CoP Inside DZ}

When the CoP is detected inside the DZ, the admittance controller implements the following law for both the strategies developed:
\begin{equation}
    \label{eq:admittance_controller_free}
    \small
    \boldsymbol{\dot{X}}_d(s) = \frac{\boldsymbol{\hat{\Lambda}}_h(s)}{\boldsymbol{M}_{adm}s+\boldsymbol{D}_{adm}},
\end{equation}
obtained from equation~\eqref{eq:admittance_controller} by setting null stiffness. Thus, the robot passively follows the humans when their CoP is inside the DZ, without hindering their movements. Mass and damping are experimentally chosen to guarantee a trade-off between transparency and stability, and they are set equal in all directions.

\subsection{Unstable state: CoP Outside DZ}
\label{subsec:unstable_state}

A graphical description of the two proposed strategies is illustrated in Fig.~\ref{fig:methods}, where Fig.~\ref{fig:methods}a shows a particular case in which a human is losing balance in the forward direction, and Fig.~\ref{fig:methods}b and c refer to the two strategies that are explained in detail below.

Henceforth, we assume that condition \{1\} occurs when the CoP is located at the border of the DZ, whereas condition \{2\} for the case in which the CoP is located outside of the DZ.
Furthermore, the matrices $\boldsymbol{M}_{\textrm{adm}}$, $\boldsymbol{D}_{\textrm{adm}}$ and $\boldsymbol{K}_{\textrm{adm}}$ of equation~\eqref{eq:admittance_controller} are computed in the admittance controller by rotating mass $\boldsymbol{M}_p$, damping $\boldsymbol{D}_p$ and stiffness $\boldsymbol{K}_p$ of a principal coordinate system, identified by directions $\boldsymbol{p}_1$, $\boldsymbol{p}_2$ and $\boldsymbol{p}_3 \in \mathbb{R}^3$, to the world frame. For instance, $\boldsymbol{M}_{\textrm{adm}}$ is computed as
$
    \boldsymbol{M}_{\textrm{adm}} = \boldsymbol{H}_{WP} \boldsymbol{M}_p \boldsymbol{H}_{WP}^T,
$ 
where $\boldsymbol{H}_{WP} \in \mathbb{R}^{6\times 6}$ is
$
    \boldsymbol{H}_{WP} = 
    \textrm{diag}\left(\boldsymbol{R}_{WP}, \boldsymbol{R}_{WP} \right)
$, 
and $\boldsymbol{R}_{WP} = \begin{bmatrix} \boldsymbol{p}_1 & \boldsymbol{p}_2 & \boldsymbol{p}_3 \end{bmatrix} \in \mathbb{R}^{3\times 3}$ is the rotation matrix from principal to world frame. The choice of $\boldsymbol{p}_1$, $\boldsymbol{p}_2$, $\boldsymbol{p}_3$, $\boldsymbol{M}_p$, $\boldsymbol{D}_p$ and $\boldsymbol{K}_p$ is driven by the specific assistive strategy as explained below.

\subsubsection{Fixed Spring Assistance (FSA)} 
In this strategy, the robot smoothly stiffens its whole-body when the CoP goes out of the DZ (Fig.~\ref{fig:methods}b), giving fixed support at the human arm level. Let $\boldsymbol{x}^*$ be the pose of the end-effector at condition \{1\}. Hence, the strategy sets 
$
\label{eq:fix_ref}
    \boldsymbol{x}_{\textrm{ref}} = \boldsymbol{x}^*.
$ 
Then, in condition \{2\} the strategy computes the first principal direction as
$
    \boldsymbol{p}_1 = \frac{\boldsymbol{r} - \boldsymbol{r}_{\textrm{ref}}}{\norm{\boldsymbol{r} - \boldsymbol{r}_{\textrm{ref}}}},
$
where $\boldsymbol{r}$ and $\boldsymbol{r}_{\textrm{ref}}$ are the end-effector and reference positions, respectively. 
Next, $\boldsymbol{p}_2$ is a randomly-chosen unitary vector orthogonal to $\boldsymbol{p}_1$, $\boldsymbol{p}_3 = \boldsymbol{p}_1 \times \boldsymbol{p}_2$ and \mbox{$\boldsymbol{K}_p = \textrm{diag}\{k_{p_1},\boldsymbol{0}_5\}$}. The parameter $k_{p_1}$ is selected based on a heuristic, in order to obtain a trade off between compliance and rigidity when a human uses the robot as a support.
Mass and damping are kept the same as inside the DZ in all directions, except for the damping along the first principal direction ($d_{p_1}$) that is modified in order to have a critically damped system, 
$
    d_{p_1} = 2\sqrt{k_{p_1}m_{p_1}}.
$

The design choices above let the robot render a virtual environment (Fig.~\ref{fig:methods}b), where a virtual spring is attached at its end-effector and grounded to the human hand position in the last equilibrium state. In this way, a force smoothly appears and attracts the humans hand towards a former stable state, preventing them from going further away.

\subsubsection{Mirrored Balance Assistance (MBA)}
On the other hand, in the second strategy (Fig.~\ref{fig:methods}c) the robot aims at bringing the human back to the upright posture (i.e., to bring the CoP inside the DZ) by applying a force at the human arm level. The computation of the direction of the desired movement uses the relative position between CoP and SP.
In condition \{1\}, the border where the CoP is located is set as the mirror line. This line is used at condition \{2\} for computing $\boldsymbol{c}_{p_m}$, the mirrored CoP. The distance from $\boldsymbol{c}_{p}$ to the mirror line is defined as $\Delta\boldsymbol{c}_p$. Therefore, $\boldsymbol{x}_{\textrm{ref}}$ is computed as:
\begin{equation}
\small
    \boldsymbol{x}_{\textrm{ref}} = \begin{bmatrix} \boldsymbol{r}_{ref} \\ \boldsymbol{\theta}_{ref} \end{bmatrix} = \begin{bmatrix} \boldsymbol{r} + 2 \Delta\boldsymbol{c}_p \boldsymbol{p}_1 \\ \boldsymbol{\theta} \end{bmatrix} ,
\end{equation}
where $\boldsymbol{\theta}$ is the current orientation of the end-effector and $\boldsymbol{p}_1$
is defined as
$
\small
    \boldsymbol{p}_1 = \frac{\boldsymbol{c}_{p_m}-\boldsymbol{c}_p}{\norm{\boldsymbol{c}_{p_m}-\boldsymbol{c}_p}}.
$
The remaining parameters are computed as in FSA.

The choices that characterize this strategy allow rendering a virtual environment in which a force proportional to the distance of the CoP from the DZ smoothly appears/disappears when the human is losing/regaining balance.

\begin{figure*}
    \centering
    \includegraphics[width=0.85\textwidth]{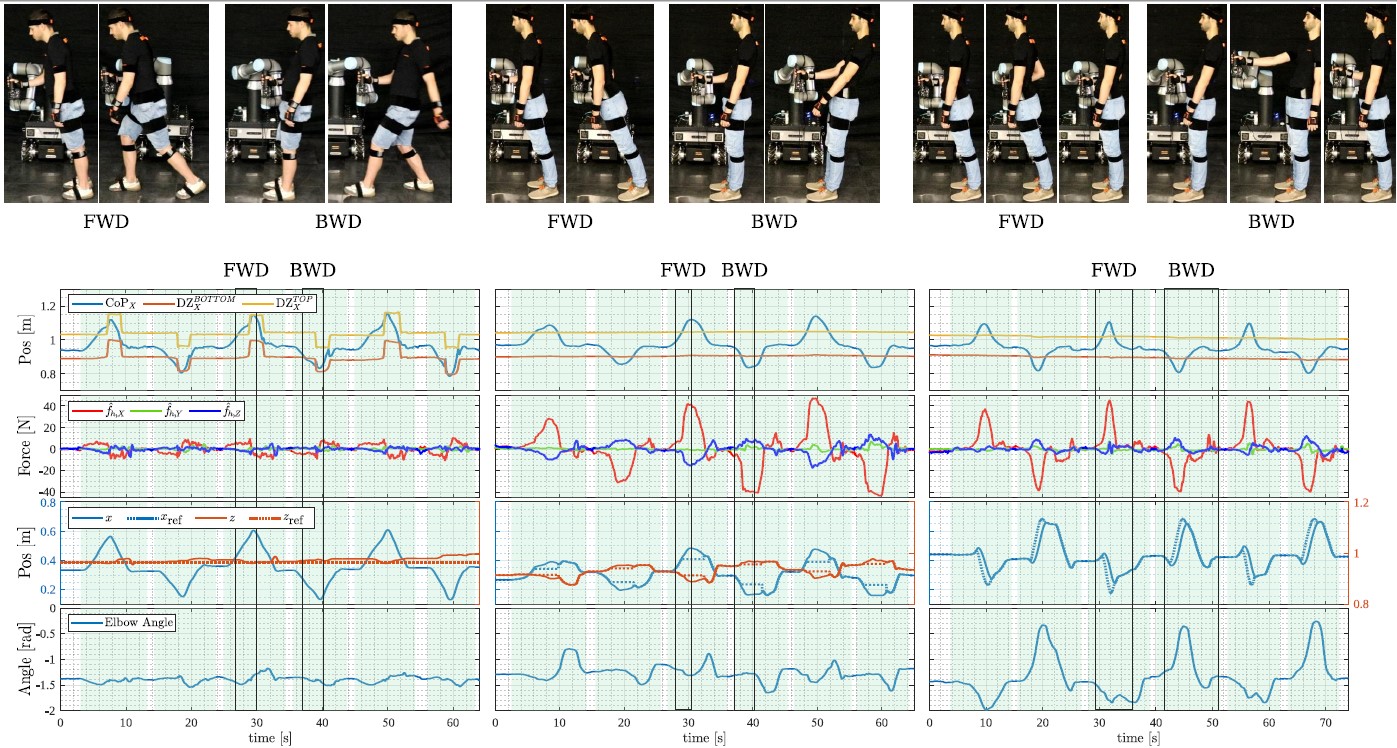}\\
    \hspace{0.2cm} (a) \hspace{4.3cm} (b) \hspace{4.3cm} (c)
    \vspace{-0.2cm}
    \caption{Quantitative results of the experiments for three different subjects obtained with HWA (a), FSA (b) and MBA (c), along with the relevant excerpts. The shaded areas represent the duration of each backward (BWD) or forward (FWD) fall, and in each graph the following quantities are plotted from top to bottom: CoP and upper and lower limits of the DZ along the X-axis, human-robot interaction forces, end-effector and reference positions in the sagittal plane (X-Z) and human elbow angle ($0~rad$ for arm completely extended, $-3.14~rad$ for arm completely flexed).}
    \vspace{-0.5cm}
    \label{fig:plot_graph}
\end{figure*}

\section{Experiments and Results}
\label{sec:experiments_results}

An experimental campaign is conducted to compare the proposed assistive strategies described in Sec.~\ref{sec:methods} along with the one presented in~\cite{XING2021102497}  (henceforth, referred to as Horizontal Wall Assistance (HWA)), which is used as the baseline. The experiments of this work were carried out considering the standing case only. This decision was taken due to the difficulty of making healthy people lose balance in a similar way to patients with neurological diseases when walking.
A video~\footnote{\url{https://www.youtube.com/watch?v=zPE_YKJCy-s}} is included as supplementary material.

\subsection{Experimental Protocol}
\label{subsec:experimental_protocol}

The whole experimental campaing was carried out at the Human-Robot Interfaces and Physical Interaction (HRII) Lab, Istituto Italiano di Tecnologia (IIT). The protocol was approved by the ethics committee Azienda Sanitaria Locale Genovese N.3 (Protocol IIT\_HRII\_ERGOLEAN 156/2020).
Twelve healthy volunteers, six males and six females, (age: $29.3 \pm 4.4$ years; mass: $65.5 \pm 13.2$ kg; height: $170.7 \pm 9.4$ cm)\footnote{Subject data is reported as: mean $\pm$ standard deviation.} were recruited. After explaining the experimental procedure, written informed consent was obtained, and a numerical ID was assigned to anonymize the data.

Prior to the experiments, the participants were asked to perform the widest oscillations possible around their upright posture without losing balance. Based on that, the DZ boundaries were computed.
Three sessions were carried out, one for each strategy. The order of the sessions was randomized to avoid learning effect.
Before each session, a familiarization phase was conducted until the subjects felt at their ease with the system and the selected strategy. Then, in each session, they were asked to lose balance along the sagittal plane, progressively bending the ankles and keeping the feet on the ground and the body straight while grasping the handle located at the end-effector of the robot (Fig.~\ref{fig:methods}a). Within each session,  the participants were required to perform six trials losing voluntarily the balance, alternating forward (FWD) and backward (BWD) with a break in between. 
Besides, the following guidelines were given:

\begin{itemize}
\item Let the robot help you to recover the balance.
\item If the robot stops providing assistance before recovering the balance, apply a force to finish the task.
\item If the robot is not providing any assistance, take a step to avoid falling.
\end{itemize}

\subsection{Assessment tools}
\label{subsec:indexes_and_metrics}

After the experiments, the participants filled a questionnaire 
to rate different qualitative aspects of their experience. The questionnaire is composed of a standard part (Trust in Automation Scale~\cite{trustQuestionnaire}) and a custom part designed specifically for this study, that includes 10 questions. Q.1 The system was intuitive to use; Q.2 I was able to recover the balance easily; Q.3 I think the robot will help me when I need it; Q.4 The system was difficult to get familiar with; Q.5 I had to make a significant effort to recover the balance; Q.6 The robot can be trusted when losing the balance; Q.7 I would use the system if I needed help to recover the balance; Q.8 I think the robot is useful for people with balancing problems; Q.9 I think it's a good idea to use the robot as balancing assistant; Q.10 I envision it as a rehabilitation system.
Besides, quantitative results are analyzed according to the following \textit{performance indexes}:

\begin{itemize}
    \item \textbf{Total time outside DZ: $ \Delta t_{out,DZ} = t_{in} - t_{out}$},  
where $t_{out}$ and $t_{in}$ refer to the instants when the CoP goes outside and inside the DZ, respectively.
If the human takes a step to recover the balance, the strategy fails, and the value of the index is not determined.
\vspace{1mm}
\item \textbf{Maximum distance from DZ:} 
\begin{equation}
\small
    D_{max,DZ} = \max_{[t_{out},t_{in}]} d_{DZ}(\boldsymbol{c}_p),
\end{equation}
where the operator $d_{DZ}(\cdot)$ is the Euclidean distance of a point from the closest boundary of the DZ. If the strategy fails the index is not determined.
\vspace{1mm}
\item \textbf{Maximum normalized force:} 
\begin{equation}
\small
    F_{max,i} = \frac{\max_{[t_{out},t_{in}]} |\hat{f}_{h,i}|}{w_h}\times100,
\end{equation}
where $w_h$ is the subject weight, $\hat{f}_{h,i}$ is the i-th component ($X,Y,Z$) of the human force. If the strategy fails (at instant $t_{fail}$) the maximum is taken over the interval $\begin{bmatrix} t_{out},t_{fail} \end{bmatrix}$.  
Since the experiments are performed inside the sagittal plane, the force perpendicular to this plane ($i = Y$) is not considered for this index.
\end{itemize}

\subsection{Results}
\label{subsec:results}
Fig.~\ref{fig:plot_graph} exhibits the relevant plots recorded during the experiments for three particular participants, one for each strategy, i.e. (a) for HWA , (b) for FSA  and (c) for MBA.  
From top to bottom, the graph shows: 
the human CoP and boundaries of the DZ along the X-axis, the three components of the force exerted by the subject, the robot end-effector and reference positions in the sagittal plane, and the human elbow angle, where a decrement corresponds to a flexion while an increment corresponds to an extension. 
Each trial is highlighted in green.

Table~\ref{tab:quantitative_results} reports the results of the quantitative analysis according to the \textit{performance indexes} for FWD and BWD falls with each strategy. All the reported data are significantly different according to sign-tests (p-value $< 0.05$).  
$\Delta t_{out,DZ}$ and $D_{max,DZ}$ are not computed for HWA since the experiment failed for both FWD and BWD. The values obtained for FSA are, on average, higher than the ones obtained for MBA. Note that these results are consistent with the plots of the three subjects showed in Fig.~\ref{fig:plot_graph}. 

 \begin{center}
\begin{table}[b]
\vspace{-0.3cm}
\caption{Results of performance indexes
for the forward (FWD) and backward (BWD) cases with each strategy (HWA, FSA, MBA).}
\label{tab:quantitative_results}
\centering
    \resizebox{\columnwidth}{!}{
    \begin{tabular}{c|c|c|c|c|c|c|c|c|c}

 \multicolumn{2}{c}{} & \multicolumn{2}{|c|}{$\Delta t_{out,DZ}$ [s]} & \multicolumn{2}{|c|}{$D_{max,DZ}$ [cm]} & \multicolumn{2}{|c|}{$F_{max,X}$ [\%]} & \multicolumn{2}{|c}{$F_{max,Z}$ [\%]} \\
 \cline{3-10}
 \multicolumn{2}{c|}{} & mean & std & mean & std & mean & std & mean & std  \\
 \hline
 \multirow{3}{*}{\rotatebox[origin=c]{90}{FWD}} & HWA & - & - & - & - & 2.74 & 1.34 & 2.69 & 1.94 \\
 & FSA & 3.73 & 1.74 & 7.61 & 3.05 & 5.92 & 1.93 & 1.65 & 1.22 \\
 & MBA & 1.89 & 0.72 & 4.14 & 1.84 & 4.18 & 1.43 & 0.37 & 0.27 \\
 \hline
 \multirow{3}{*}{\rotatebox[origin=c]{90}{BWD}} & HWA & - & - & - & - & 2.46 & 1.52 & 0.70 & 0.43 \\
 & FSA & 3.41 & 1.93 & 5.89 & 3.18 & 4.87 & 1.71 & 1.83 & 1.38 \\
 & MBA & 2.37 & 1.31 & 4.31 & 2.36 & 4.25 & 1.80 & 0.99 & 0.47 \\
    \end{tabular}
   }
\end{table}
\end{center}

\vspace{-4mm}
Concerning the maximum normalized forces, for HWA the results are in agreement with Fig.~\ref{fig:plot_graph}a. In BWD, $F_{max,X} > F_{max,Z}$ due to the high horizontal speed reached right before taking a step and due to the upward movement of the human hand. Instead, in FWD, $F_{max,X}$ and $F_{max,Z}$ are comparable since the subjects move the hand downward, pushing slightly against the horizontal wall. In FSA, the indexes take on higher values than MBA due to the higher value of $D_{max,DZ}$. In this case, participants apply a force also along the X-axis. Finally, for MBA, $F_{max,X} > F_{max,Z}$, since this strategy applies a force on the human arm along the X-axis. 

\begin{figure}
    \centering
    \includegraphics[width=0.9\columnwidth]{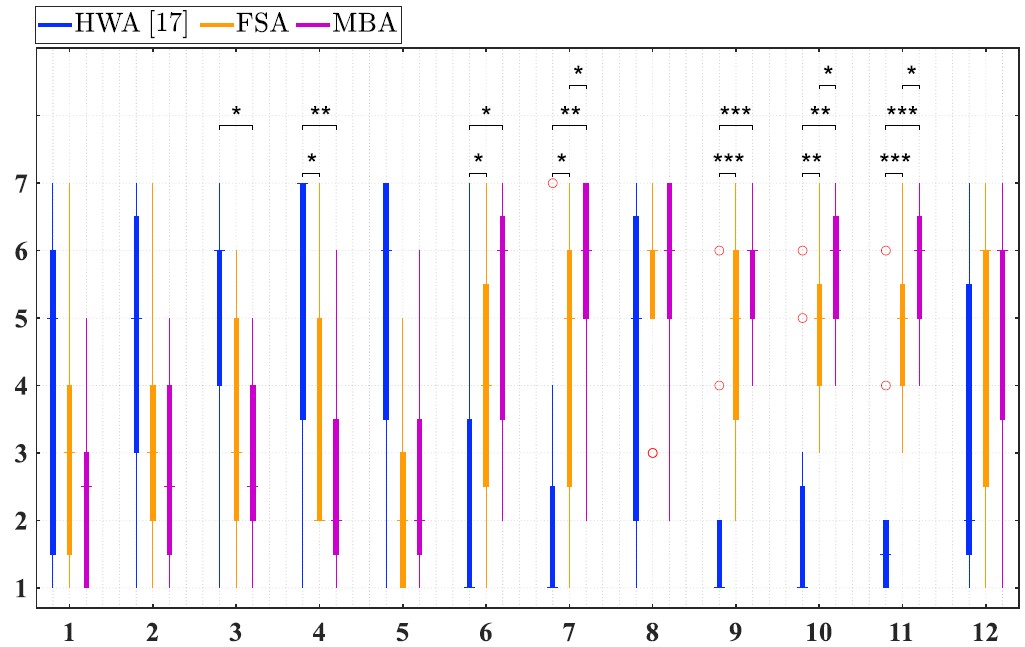}\\
    (a)\\ \vspace{0.3cm}
    \includegraphics[width=0.9\columnwidth]{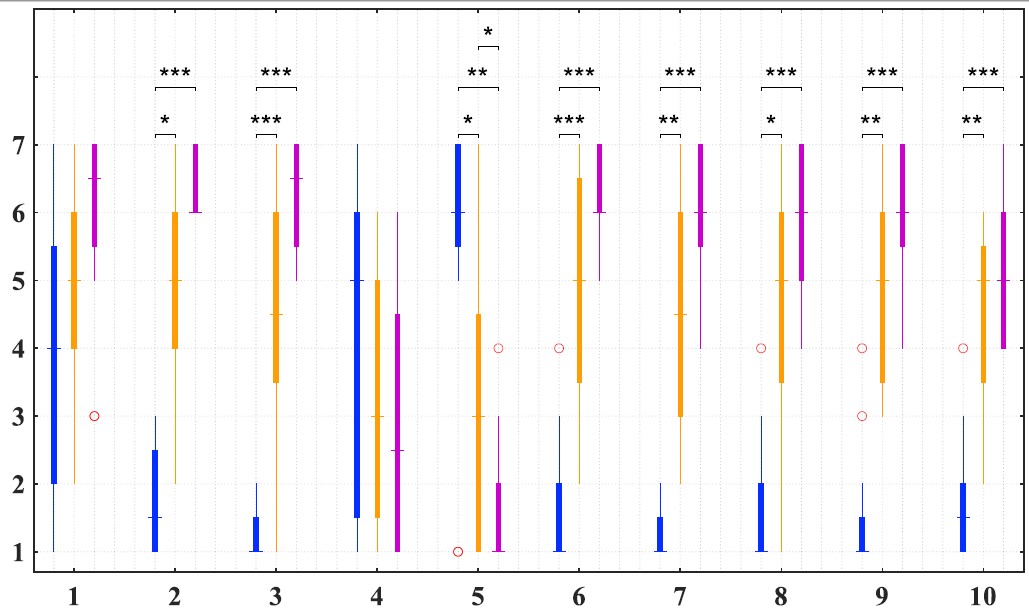}\\
    (b)
    \caption{Results obtained from the standard (a) and custom (b) questionnaires for the three strategies along with the outcomes of the statistical test carried out for each pair of strategies: *: $p<0.05$, **: $p<0.01$, ***: $p<0.001$, nothing: not significant.}
    \vspace{-0.5cm}
    \label{fig:qualitative_results}
\end{figure}

Fig.~\ref{fig:qualitative_results} depicts the findings of questionnaires. For each questionnaire statement, the results of the three strategies are represented as box-plots and the outcomes of the sign-tests conducted for each pair of strategies are reported.

Focusing on the standard questionnaire (Fig.~\ref{fig:qualitative_results}a), the strategies show significant differences for statements 7 (security), 10 (reliability), and 11 (trust), where the scores improve from HWA to MBA. For assertion 3 (suspicion), only HWA and MBA are significantly different, with HWA having a worse evaluation than MBA. Finally, HWA is significantly different from FSA and MBA for statements 4 (skepticism), 6 (confidence), and 9 (trust), where the last ones get superior scores than the first one. Nevertheless, no significant difference is noticeable for statements related to unexpected or dangerous robot behaviors (1, 2, and 5), system integrity (8), and familiarity with the system (12).

In the custom questionnaire (Fig.~\ref{fig:qualitative_results}b), statement 5 (performance) features significant differences for all the pairs of strategies, and the scores improve from HWA to MBA. For statements, 7, 8, and 9 (usability), 10 (applicability), 3 and 6 (trust), and 2 (performance), HWA is significantly different from FSA and MBA, which are rated better than the former, while the differences between FSA and MBA are not significant.
On the other hand, statements 1 (intuitiveness) and 4 (familiarity with the system) do not show any statistically significant difference.

\section{Discussion}
\label{sec:discussion}

Based on the quantitative and qualitative results previously presented, the following observations can be remarked.
HWA is the only strategy that fails. 
Only one participant was able to recover the balance for the FWD case. A possible explanation lies in the force components that the strategies provide to assist the user.
Even though HWA offers a sustain along Z, most of the subjects did not use it. This fact suggests that the Z component of the force is less intuitive to recover the balance in the setup used. 

Overall, results show that FSA and MBA outperform HWA. In particular, 
the subjects evaluated HWA as less usable, trustful, applicable, and performing than FSA and MBA. Moreover, they were also skeptical, less confident, and suspicious of the system when HWA was used.

When using the FSA strategy, after staying for a while outside the DZ, the subjects voluntarily applied a force either flexing (BWD) or extending (FWD) the elbow in order to move the CoP back inside the DZ.
On the contrary, by applying the MBA strategy the CoP is immediately brought back inside the DZ. In particular, the robot is now flexing (FWD) or extending (BWD) the human elbow without requiring the subjects to apply any force. This difference between FSA and MBA can be appreciated in the elbow angle graphs, that are $180^{\circ}$ out of phase. 
Indeed, FSA, similarly to HWA, only constraints the humans' motion without proactively helping them regain the upright posture.
This aspect probably affected the responses in the questionnaire, where MBA obtained a higher rating than FSA for statements concerning performance, security, reliability, and trust.
Moreover, MBA allowed lower deviations of the CoP from the DZ. Thus, the maximum force applied is smaller in MBA than in FSA, resulting in higher efficiency when MBA is used.

Interestingly, some subjects preferred FSA in terms of performance, security, reliability, and trust, which explains why albeit MBA scored better than FSA for those parameters, the difference is not highly significant from a statistical point of view (\mbox{$0.01<p<0.05$}). This might be due to the use of healthy subjects as they only need a sustain that FSA sufficiently provides since they can rely on their force to regain balance.

\section{Conclusions}
\label{sec:conclusions}

This paper proposed two strategies for human balance assistance using a mobile collaborative manipulator. These two approaches were integrated into an interaction-assistive whole-body framework that leveraged human state feedback to provide the necessary physical assistance. 
Based on CoM and SP information, these approaches detect loss of balance and act accordingly to compensate for such situations. The two proposed approaches were experimentally evaluated on healthy human participants in a standing situation. A quantitative and qualitative analysis was carried out to evaluate the performance of both strategies. Moreover, a comparison of our proposals with the strategy previously proposed in~\cite{XING2021102497} is conducted. The outcomes of this work stated that our strategies perform better than the baseline methodology. In this respect, one conceived approach (MBA) manifested a significantly superior performance than the other (FSA).

Future work will focus on extending our approaches to the walking case. Besides, as our strategies are intended for the elderly and people with neurological disorders, a future research line will involve patients with actual equilibrium problems to evaluate our proposal in a real scenario. Moreover, the integration of various indexes will be studied to enhance the balance loss detection both for the walking and real-scenario cases.

\bibliographystyle{IEEEtran.bst}
\bibliography{biblio.bib}

\end{document}